\documentclass[conference]{IEEEtran}
\IEEEoverridecommandlockouts
\usepackage{cite}
\usepackage{amsmath,amssymb,amsfonts}
\usepackage{algorithmic}
\usepackage{graphicx}
\usepackage{textcomp}
\usepackage{xcolor}
\usepackage{hyperref}
\def\BibTeX{{\rm B\kern-.05em{\sc i\kern-.025em b}\kern-.08em
    T\kern-.1667em\lower.7ex\hbox{E}\kern-.125emX}}

\newcommand{\mrmp}{{\textit{MRMP }}}

\begin{document}

\title{A Survey of Multi-Robot Motion Planning
}

\author{\IEEEauthorblockN{Hoang-Dung Bui}
\IEEEauthorblockA{\textit{Computer Science Department } \\
\textit{George Mason University},\\
Fairfax, VA - 22030 \\
hbui20@gmu.edu}}


\maketitle

\begin{abstract}
Multi-robot Motion Planning (MRMP) is an active research field which has gained attention over the years. \mrmp has significant roles to improve efficiency and reliability of multi-robot system in a wide range of applications from delivery robots to collaborative assembly lines. This survey provides an overview of \mrmp taxonomy, state-of-the-art algorithms, and approaches which have been developed for multi-robot systems. Moreover, it discusses strengths and limitations, and applications of the algorithms in various scenarios. From the discussion, we can draw out open problems which can be extended in future research.
\end{abstract}


\section{Introduction} \label{sec:intro}

\begin{figure}[t]
    \centering
    \includegraphics[scale=0.17]{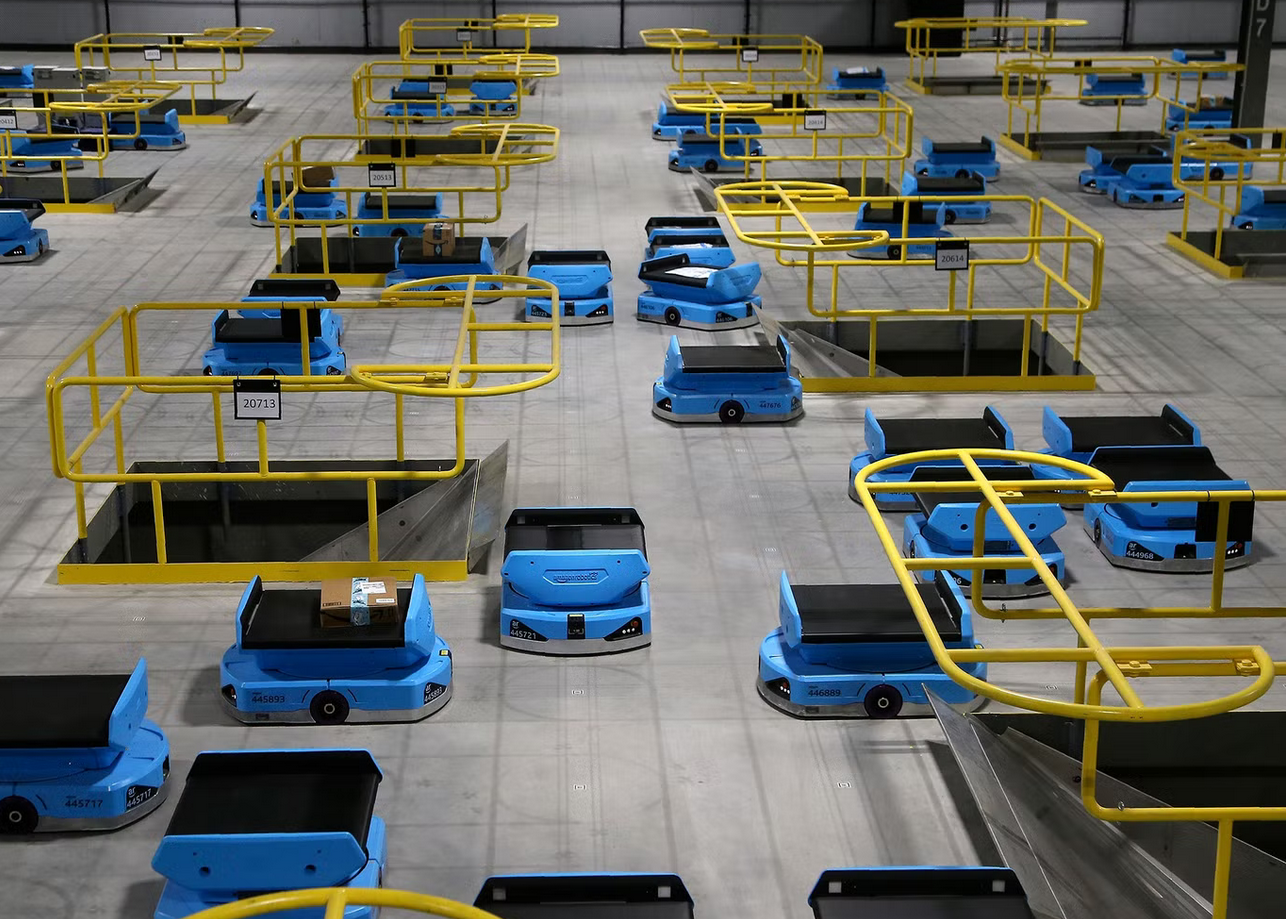}
    \caption{A multi-robot team in Amazon warehouse \href{https://www.independent.co.uk/news/business/robots-amazon-delivery-artificial-intelligence-technology-a9264036.html}{(source)}}
    \label{fig:Industry}
\end{figure}

Multi-robot motion planning (MRMP) is a critical area of research in robotic fields that involves the coordination and control of multiple robots to move from initial positions to goal locations without colliding with obstacles or each other. 
As applications of robot are increased in various industries, multi-robot systems are becoming more prevalent in diverse fields, such as warehousing \cite{b1}, multi-robot teams \cite{b2}, aircraft management \cite{b3}, and digital entertainment \cite{b4}. 
Fig.~\ref{fig:Industry} shows an example of a multi-robot team being used to pickup and retrieval products in an Amazon warehouse.

Major challenges in MRMP are to develop efficient algorithms and techniques that can handle numerous robots operating in complex, unstructured, obstacle-rich environments while ensuring optimal performance and safety. They require integration of various disciplines, including artificial intelligence, machine learning, and robotics. In the context, this survey elaborates fundamental concepts, challenges, current advancements, and open problems in MRMP.

The paper is divided into several sections. In Section \ref{sec:prelimi}, we cover the basic and general concepts of {MRMP}. Section \ref{sec:taxonomy} presents a taxonomy that classifies multi-robot motion planning based on tasks, motion types, environments, and information sharing among robots. Section \ref{sec:survey} examines different classes of {MRMP}, including state-of-the-art algorithms, and explores their advantages and limitations. In Section \ref{sec:openpro}, we identify several open problems in multi-robot motion planning that still require further research. Finally, Section \ref{sec:con} provides concluding remarks and general observations.

\section{Preliminaries} \label{sec:prelimi}
Multi-robot motion planning refers to coordinating motion of multiple robots in order to reach a specific goal or objective. This can be challenging because each robot has its own objectives, dynamics, geometries, and limitations that must be considered in planning processes. There are various types of MRMP problems, such as cooperative motion planning, and competitive motion planning. In the former one, the robots work together to achieve a common goal, while in the latter one, the robots may be competing with each other for resources or objectives.

Finding optimal solutions for \mrmp problems is NP-hard \cite{b15} and there are various methods to deal with it such as centralized planning, decentralized planning, and prioritized planning. Centralized planning involves a central planner that coordinates motion of all robots. In decentralized planning, each robot independently makes its own decisions based on local information; their paths are then collision checked and replan if a conflict occurs. Prioritized planning assigns a priority to each robot, then planning from the highest priority robot to the lowest one. The paths of the lower priority one must not violate the paths of the higher one. 
Algorithms in the approaches are compared in several criteria such as completeness, coverage of free space, optimality, and complexity analysis.  

Defining a \mrmp problem depends on robot's models and their features. If the robot's models are considered as points or simple shapes with no dynamics, the \mrmp will work in workspace or configuration space $\mathcal{W}$ which are then represented as graphs. There are $n$ agents/robots at initial positions $s$ attempting to reach their goals $t$ in $\mathcal{W}$. Each agent takes one action at a time, and the action can be $move$ or $wait$. A \textit{valid solution} to a \textit{MRMP} problem is a joint plan, in which the agents follow their corresponding plans to move from the starting positions to the goals without collision with obstacles or other agents. A collision occurs if two agents share a common vertex or moving on the same edge at the same time step, or agents go inside obstacles.

Considering dynamics and differential constraints, each robot model is defined a tuple $\mathcal{R}_i = <\mathcal{P}_i,\mathcal{S}_i,\mathcal{A}_i, f_i>$, in terms of its geometric shape $\mathcal{P}_i$, state space $\mathcal{S}_i$, action space $\mathcal{A}_i$, and motion equation $f_i$ \cite{b10}. The state space $\mathcal{S_i}$ is represented by a finite set of continuous variables, which usually consists of positions, orientation, steering angle, and velocity. To move the robot, we need to apply control action on it such as acceleration and turning the steering wheel. The actions space $\mathcal{A_i}$ is defined as a set of all the control actions that can apply on the robots. The control values are usually bounded. The motion equation $f_i$ encapsulate the underlying robot dynamics which describe the relation between the robot's state and control action. $f_i$ is often expressed as a set of differential equations of the form:
\begin{align}
    \dot{s} = f_i(s,a)
\end{align}
where $s \in \mathcal{S}_i, a \in \mathcal{A}_i$, and $\dot{s}$ is the derivative of $s$. The robot's models allow non-holonomic constraints and nonlinear equations with first, second, or higher order derivatives. A trajectory $\zeta_i:[0,T] \rightarrow \mathcal{S}_i$ is a continuous function with $T\in R^{\geq 0}$ is a duration. The trajectory $\zeta_i$ is dynamically feasible if it meets the differential constraints imposed by equations $f_i$. Collisions are implemented via two checks: between robot $i$ with shape $\mathcal{P}_i$ and robot $j$ with shape $\mathcal{P}_j$ at state $s$ at time $t$, and between a robot $i$ and obstacles. A dynamically feasible trajectory is collision-free, if there is no collision as the robot moving along it. We expect that a \mrmp solver will return a set of dynamically feasible and collision-free trajectories if it exists.   

There are two common measures to evaluate a \mrmp solution: \textit{makespan} and \textit{sum-of-costs} (SOC) \cite{b14}. The \textit{makespan} of a joint plan $\Pi$ denoted $M(\pi)$ is the time period until the last agent reaches its goal.
\begin{align*}
    M(\Pi) = \underset{1 \leq i \leq k}{max} \lvert \pi_i \rvert
\end{align*}
The sum-of-cost of a joint plan $\Pi$ - $SOC(\Pi)$ is the sum of actions or time until all agents reach their targets.
\begin{align*}
    SOC(\Pi) = \underset{1 \leq i \leq k}{\sum} \lvert \pi_i \rvert
\end{align*}

In this survey, we assume that a robot waits at its goal also increasing the overall joint plan \textit{SOC} unless it does not plan to move later from that goal.

\begin{figure}[t]
    \centering
    \includegraphics[scale=0.26]{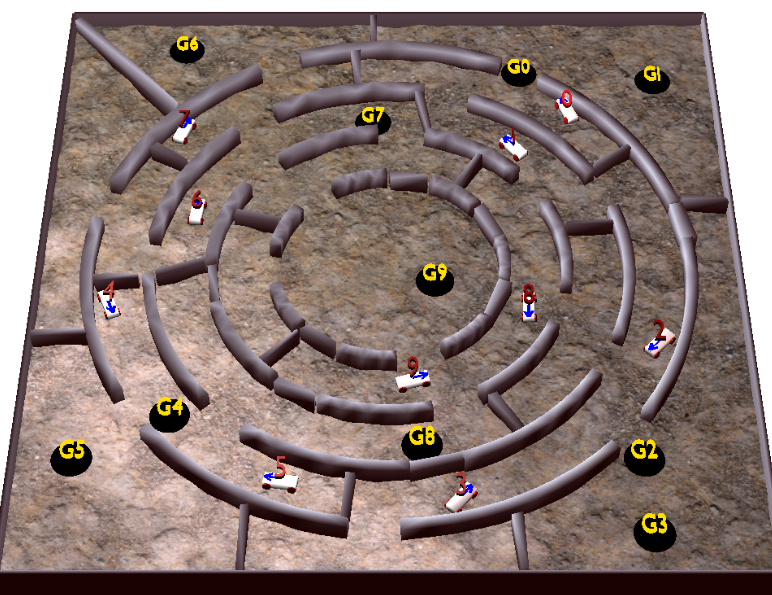}
    \caption{\textbf{An example of a multi-robot motion-planning problem}. Each robot must reach its goal (robot $i$ should go to goal $G_i$), while avoiding collisions with obstacles (shown in gray) and each robot.}
    \label{fig:preliminary1}
\end{figure}



\section{Area Taxonomy} \label{sec:taxonomy}
\begin{figure*}[htbp]
    \centering
    \includegraphics[scale=0.45]{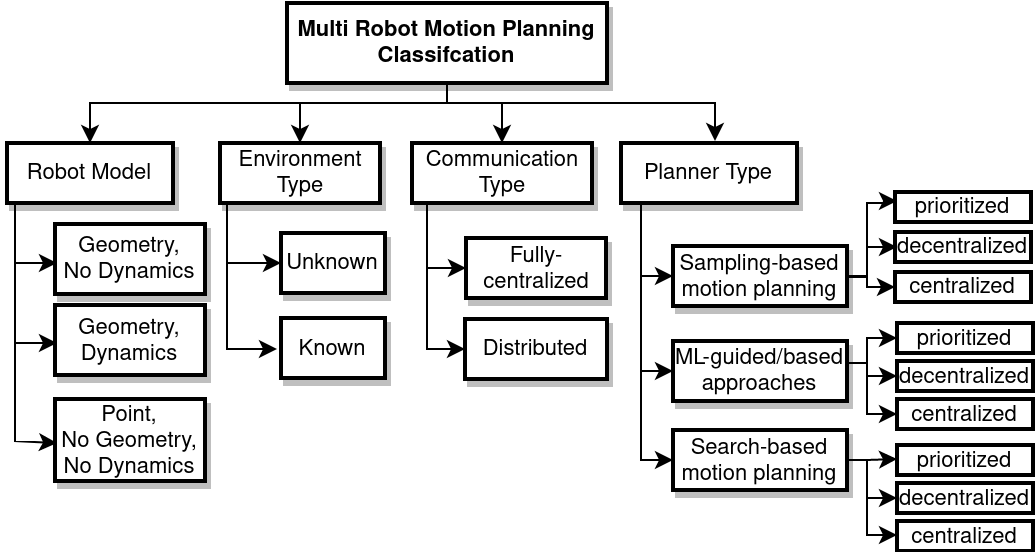}
    \caption{Taxonomy in Multi Robots Motion Planning}
    \label{fig:taxonomy}
\end{figure*}

In this section, we provide a high-level explanation of a \textit{MRMP}'s taxonomy from robotics and artificial intelligence perspectives, with focusing on single objective motion plannings. Algorithms in each class will be discussed regarding their computation time, completeness, and optimality. Considering features such as robot models, environment type, communication type and planner type, \textit{MRMP} problems can be classified as shown in Fig. ~\ref{fig:taxonomy}.

\mrmp's problems can be classified into (1) \mrmp based on Robot Models, (2) \mrmp based on Environment Type, (3) \mrmp based on Communication Type, and (4) \mrmp based on Planner Type. In the first class, Robot Model-based algorithms are grouped into three subclasses whose Robot Models are considered: (1) points with no geometry and no dynamics, (2) geometry and no dynamics, and (3) geometry and dynamics. In the second class, Environment-based algorithms are divided into two groups: Working in Complete Known Environment and Unknown Environment. In known environments, workspace information is provided completely to all robots. In unknown ones, robots use sensors to discover workspace and update discovered data on their maps. 
In the third class, we classify Communication-based algorithms into Fully-centralized approaches and decentralized approaches. In the fully-centralized approaches, all data are shared among robots synchronously. In the decentralized setting, the robots only communicate and share their data with a group of neighboring robots. In the last class, Planner-based algorithms are put into Search-based motion planning, Sample-based motion planning, and machine learning (ML)-guided/based approaches. Search-based planning relies on graphs to compute paths for trajectories over a discrete representation of the problem. Sampling-based planning is a method that does not explore fully the configuration space, which significantly increases the planning performance. \textit{ML}-guided/based approaches leverage the significant progress of \textit{ML} to improve the performance of motion planners. 
The following section will elaborate these classes.


\section{Taxonomy-Based Survey}\label{sec:survey}
This section discusses algorithms in each class, which is shown in Fig. \ref{fig:taxonomy}. We presented the algorithms in order of fast algorithms (not complete or optimal), to complete ones, then optimal ones. Several motion planners with learning are also mentioned in each class. 

\subsection{MRMP Based on Planner Type}
From the planner type's perspective, \mrmp algorithms can be classified into (1) Sampling-based motion planning, (2) Search-based motion planning, and (3) Machine Learning guided/based motion planning. 

\subsubsection{Search-based Motion Planning} \label{sec:search_base}
Search-based planning - termed Multi-agent Path Finding \textit{MAPF} - relies on graphs to compute paths or trajectories over a discrete representation of the problem. Time is discretized into time steps, and each agent takes place at a vertex. A collision occurs if two agents occupy one vertex or travel on a same edge. Two main problems in the search-based method are (1) how to turn a problem into a graph, and (2) how to search the graph to find the best solution. Various search-based algorithms applied to \textit{MRMP} are proposed in \cite{b5, b19, b20, b21,b22, b24, b26, b27,b28, b30, b31, b54, b55, b56}.  

\textit{Windowed Hierarchy Cooperative} $A^*$ ($WHCA^*$) \cite{b54} implements a cooperative search, using hierarchical heuristics to guide an order in which agents should be selected. Conflicts among agent's paths are limited to a window of the next $m$ time steps (not the entire trajectory).
\textit{Push-and-Swap} \cite{b19} and \textit{Push-and-Rotate} \cite{b20} are complete algorithms which solve efficiently up to $\vert V \vert -2$ agents in the graph $G=(V,E)$. \textit{Push-and-Swap} introduces two rules: \textit{push} an agent toward its goal if the next vertex is non-occupied, and \textit{swap} - allows two robots to exchange their location without changing to other robots. \textit{Push-and-Rotate} separates the graph into subgraphs in which the agents can reach any vertex in the subgraphs. It uses three operations: \textit{push, swap}, and \textit{rotate} to find solutions.  
MAPP \cite{b21}, and BIBOX \cite{b22} are two complete algorithms that solve MAPF problems with \textit{slidable} feature fast and efficiently. \textit{Slidable} feature is for any triple of vertices $v_1, v_2,$ and $v_3$, there exists a path connecting $v_1$ and $v_3$ not passing through $v_2$. 
Those algorithms are fast and provide suboptimal solutions.

Several algorithms are able to solve MAPF problems completely or optimally, including the increasing-cost tree search \textit{(ICTS)} \cite{b24}, conflict-based search - \textit{CBS} \cite{b25}, $A^*$ and its variants \cite{b5, b30, b31}, $M^*$ \cite{b26}, and Constraints Programming \cite{b27,b28}. 

\noindent \textit{Extension of} $A^*$: In \cite{b5}, the authors proposed two techniques - Operator Decomposition (\textit{OD}) and Independence Detection (ID) - to speed up $A^*$ algorithm. \textit{OD} reduces a significant number of generated nodes by updating positions of a single agent only in the composite state space. $A^*$ with \textit{OD} is considered as a global search algorithm. \textit{ID} algorithm is used as a function inside the global algorithm and partitions the agents into several smaller groups and finds optimal paths for each group using \textit{OD}. \textit{ID} starts by finding paths for agents individually ignoring the other agents; then collision is checked among the paths. The two first conflict agents are grouped to find their optimal paths by using \textit{OD}. The merging of two conflict agent into a group and finding optimal paths for the merged group keep going until there is no more conflict. The running time is dominated by running \textit{OD}. The agent's group merging can be avoided if we can find another set of optimal paths for one of the agent's groups. The algorithm can provide optimal solutions.

\textit{Enhanced Partial Expansion} $A^*$ ($EPEA^*$) \cite{b30} uses the concept of \textit{surplus} nodes whose estimated costs are larger than the optimal cost $f(n) > C^*$. The \textit{surplus} nodes do not contribute to finding the optimal paths and are never to be expanded; thus, they should be pruned out to save time for generating these nodes. In \cite{b31}, the authors convert MAPF problems into network flows and introduce a \textit{flow-based heuristic} - which is a summation cost of the shortest paths which are used for all agents reaching any vertex in goal's set without collision. This heuristic is admissible because it is smaller or at most equal to the cost for all agents reaching their desired goals. Using \textit{flow-based} heuristic with \textit{OD+ID} brings an advantage over the ordinary \textit{OD+ID} algorithm. 

\noindent $M^*$ \cite{b26} is a sub-dimensional expansion framework, which deals with exponential expansion of state space by changing dynamically branching factor of search space. In this aspect, $M^*$ is similar to \textit{OD} which limits the factor of some vertices. The sub-dimensional expansion method relies on two components: \textit{individual policy} and \textit{collision set}. 
Initially, $M^*$ expands graph $G^{exp}$ from a vertex $v$ by generating a new vertex $v_{new}$ using individual agent policies to generate individual steps for their own, optimal paths. If there is no collision, the vertex expansion will continue in a one-dimensional search space. If a collision occurs between two agents $i$ and $j$, they are added into \textit{collision set}; then the collided paths are allowed to diverge from their own optimal paths. All the vertices along the paths from the start vertex $v$ are re-expanded with all possible action combinations of agents $i$ and $j$. After finding collision-free paths, $M^*$ continues running the vertex expansion. Each vertex of $M^*$ stores a \textit{conflict set} of agents for which it generates all combinations of actions. For agents not in \textit{conflict set}, they will take single actions which are on their individual optimal paths. 

\textit{Recursive} $M^*$ \cite{b26} ($rM^*$) is an enhanced version of $M^*$. $rM^*$ attempts to avoid a large action combination from all collided agents, but decoupling them into smaller groups. This feature is similar to \textit{ID} in which sets of agents can be solved separately. Practically, \textit{OD, ID} can be used in $rM^*$ framework to bring advantage in some scenarios \cite{b26}.

\noindent In $ICTS$ algorithm \cite{b24}, a \textit{complete solution} for a \mrmp problem is determined by a set of individual paths $<\sigma_1,...,\sigma_n>$ for individual agents. The planning search consists of two levels. The \textit{high-level search} - performs a search on a tree called - \textit{increasing cost tree} (ICT). Each node on \textit{ICT} consists of a vector $C = <c_1,...,c_n>$ - $n$ is the number of agents and each value $c_i$ is the cost of the path of agent $i$ to reach its goal. An $ICT$ node represents all possible solutions whose cost of the individual agent path is equal to $c_i$. For each visited $ICT$ node, the \textit{low-level search} - verifies if there is a valid solution whose cost of each path is equal to the corresponding element in vector $C$. To find the paths efficiently, $ICTS$ computes all possible plans for a single agent $i$ with cost $c_i$ by breath-first search. All those paths with the given cost and the corresponding agent, are stored in a \textit{Multi-value Decision Diagram} (MDD) data structure. $ICTS$ algorithm is sped up by a pruning technique, which quickly checks there are feasible paths for the agent in a $ICTS$'s node. If the search in \textit{MDD} returns \textit{null}, the \textit{low-level search} can safely return the solution for the corresponding $ICT$ node.

\noindent \textit{CBS} \cite{b25} is an optimal algorithm which solves a \textit{MAPF} problem by a sequence of single-agent path finding problems. Similarly to \textit{ICTS}, \textit{CBS} consists of two searches: a \textit{low-level} search and a \textit{high-level} search. Both the searches run on a Constraint Tree (\textit{CT}), whose nodes consist of joint plans for all agents and the constraints for the agents to comply. The constraints are in the form: $<i, v,t>$ meaning agent $i$ is not allowed to occupy vertex $v$ on the graph $G$ at time step $t$.
The \textit{low-level} search works on a \textit{CT}'s node and performs an optimal single-agent pathfinding to each agent (ignoring other agents) under constraints that are imposed by the \textit{high-level} search. Any optimal single-agent path finding such as $A^*$ can be used in the \textit{low-level} search. 
The \textit{high-level} search selects arbitrarily a node $v$ on \textit{CT} to process. First, it checks collision among the paths. If there is no collision, it returns  the $found\_solution$. If collisions exist, the search will set the first conflicts in the form: $<i,j,v,t>$ meaning agent $i$ and $j$ colliding at vertex $v$ at time step $t$. From this conflict, two constraints: $<i,v,t>$ and $<j,v,t>$ are created, and each is added into two new vertices $v_l$ and $v_r$. Both vertices are $v$'s children. The \textit{low-level} search chooses one of those nodes to work further. 

There are many extensions and improvements for \textit{CBS}. \textit{Meta-CBS} \cite{b25} improves efficiency by creating a meta-agent instead of constraints to resolve a conflict between two paths. This reduces the branching factor of \textit{MT}, therefore the problem dimension is much smaller to \textit{CBS}. An optimal MAPF solver can be used to handle the meta-agent which has a much smaller dimension. Improved-CBS \cite{b32} speed up \textit{CBS} by reducing the size of the \textit{CT} by using concepts of \textit{cardinal}, \textit{semi-cardinal}, and \textit{non-cardinal} to decide to split or merging the constraints. In \cite{b33}, the authors proposed a heuristic at the \textit{high-level} search to select a more potential \textit{CT}'s node to expand. In many testing scenarios, it handled well up to 40 agents with a high success rate.

\textit{ML-CBS} \cite{b44} uses \textit{ML} in \textit{CBS}'s framework to increase the quality of conflict's selection to solve. A neural network mimics an oracle behavior, which can predict conflicts that result in a smaller size of the constraint trees. \textit{ML-CBS} reduces the running time from 10.3\% to 64.4 \% over \textit{CBS} in the tested environments.



\subsubsection{Sampling-based Motion Planning} \label{subsec:samp-based}
Sampling-based planning is a method that does not explore fully the configuration space, which significantly increases planning performance. Sampling-based approaches consist of sampling randomly configuration spaces, and reasoning on a much smaller set of sampled configurations.

Multi-robot sampling-based motion planners such as \textit{RRT-based} \cite{b6,b7,b11,b12}, \textit{GUST} \cite{b8}, \textit{KPIECE} \cite{b9}, \textit{CoSMMAS} \cite{b10} approach \textit{MRMP} problems in a centralized way and building motion trees $\mathcal{T}$ to find trajectories. 
Those planners incorporate robot's geometries and dynamics to progress  motion trees' expansion. Vertices – containing robot's states – can be added to $\mathcal{T}$ if they are collision-free and allow the robots to move safely from the current state to that new state. To generate a new vertex for the motion tree, we apply control actions to the robot and numerically integrate the motion's differential equations.
Adding a new vertex by this approach helps the motion trees handle well two-boundary values problems and have advantages over roadmaps and analytical solutions, which are limited to solving the problems in some special cases or too computationally expensive, respectively.

Another problem arising with those algorithms is the motion trees' expansion easily fails if the node is near obstacles (boundary node). To avoid the near obstacles areas, the motion planners needed to be guided to areas with a low chance to get collisions. We will discuss the algorithms' metrics to guide the motion tree expansion. 

Dynamic-Domain RRT (DD-RRT) \cite{b11} handles well bottle-neck areas by controlling sampling domain of boundary nodes to a ball with a small predetermined radius. At a visibility region, the radius is set to a much larger value. Similar to \textit{DD-RRT}, Adaptive Dynamic Domain RRT (ADD-RRT) \cite{b12} proposed the sampling domain ball with adaptive radius. The radius is tuned according to the extension success rate of each boundary node. 

Kinodynamic Motion Planning by Interior–Exterior Cell Exploration (KPIECE) motion planner \cite{b9} is designed for systems with complex dynamics. The idea is to guide the planner sampling in less-explored areas in state space. The state space's coverage is estimated by a grid-based discretization, which also keeps track of the boundary of explored regions. If the area is sampled multiple times, its score will be reduced, and other areas are selected to sample.

Guided Sampling Tree (GUST) planner \cite{b7} is a dynamic sampling-based tree search that expands a motion tree $\mathcal{T}$ by using region-guided. It decomposes a workspace into non-overlapping regions and calculates heuristic paths from the region's centroids to a goal. The paths are segmented into waypoints and are used to guide $\mathcal{T}$ expanding. To map the state space into the configuration space, the authors built a projection, which extracts robot's position $(x,y)$ in the state space into the workspace. Motion tree $\mathcal{T}$ is developed by selecting a vertex $v$ with the lowest cost to the goal.  The planner selects a region that contains $project(v)$ and determines a path from the region to the goal. The robot makes a new state $v_{new}$ by applying a control action and integrating the robot motion equation $f$ for one time step $dt$ heading to the closest waypoint on the paths. The new state is projected into the workspace to check for collision. If no collision, $v_{new}$ and edge $v, v_{new}$ are added into $\mathcal{T}$, the regions, and the groups which are corresponding to each waypoint. After a number of times failing to find a trajectory, we penalize the selected region and pick another one to recalculate.  

Leverage \textit{GUST}, Cooperative Sampling-based Multi-Robot Motion Planning and Multi-Agent Search (CoSMMAS) \cite{b10} is proposed to handle multi-robot systems. There are several major components in \textit{CoSSMAS}: (1) Discrete Abstraction of an environment, (2) Multi-agent Search, (3) A Motion Tree ($T$), and (4) Equivalence Classes.

Discrete Abstractions are set up by constructing a roadmap $\mathcal{M}_i$ over each configuration space $\mathcal{C}_i$. $\mathcal{M}_i$ (ignoring the robot's dynamics) is built by sampling collision-free configuration and connecting the neighboring nodes by checking the collision-free paths. The road map is the input to the Multi-agent Search algorithm which can be any multi-agent path-finding algorithm (no dynamics here) to output a set of valid paths. The paths are used as a guide for the motion planning algorithm to build $\mathcal{T}$. The motion tree $\mathcal{T}$ is expanded in the composite state space $\mathcal{S}_1 \times ... \times \mathcal{S}_n $ to account for the robot dynamics. The procedure to generate a new state $v_{new}$ starts by picking the state $v$ whose weight is the highest. We apply some control actions $<a_1,...,a_n>$ and integrating the motion equations $<f_1,...,f_n>$ of each robot for one time step $dt$. If $v_{new}$ and the path connecting $v$ and $v_{new}$ are collision-free, we add $v_{new}$ and the edge $<v, v_{new}>$ into $T$. $v_{new}$ can be expanded in any direction uniformly, however, to speed up the expansion, we use the valid paths as the guidance. The states are mapped to the configurations in the roadmap only considering the location and orientation $(x,y, \theta)$ of the robot. If multiple states map to a configuration, they will be assigned to an equivalence class. 
Equivalence classes are convenient for the multi-agent search on the configuration space and expanding $T$ because as following along the valid paths, we just need to select vertices from the corresponding equivalence class.

There are several frameworks that combine \textit{MRMP} with \textit{ML} \cite{b46,b47}. These frameworks work well with up to five robots, however, they are underdogs to \textit{CoCMMAS} which can handle up to ten robots.

The sampling-based planners are the probabilistic complete algorithms, however, they fail to find the optimal solutions. To the best of my knowledge, there is no \mrmp planner that can provide optimal solutions when working with dynamic robot systems.


\subsubsection{ML-guided/based approaches}
Machine Learning (\textit{ML}) has significant progress recently and is applied widely in motion planning. \textit{ML} can be used to develop \mrmp solvers \cite{b46, b47} or to estimate heuristics or motion planner selector \cite{b44, b43}.

In \cite{b47}, Tingxiang al et. proposed a decentralized multi-robot collision avoidance framework in which each robot can make its own navigation decision without communication with other robots. The navigation policy is a combination of PID policy, \textit{RL} policy, and Safe policy, which will work on simple scenarios, complex scenarios, and emergent scenarios, respectively. The switching among the scenarios depends on the distances from the robots to the closest obstacles and to their targets. If the distance between a robot and its goal is closer than to any obstacle, the robot uses \textit{PID policy} to go straight to the goal. If the distance to any obstacle or other robot is less than a $r_{risk}$, the robot will use \textit{Safe policy} to move with caution (avoid collision). Otherwise, \textit{RL policy} will be deployed. \textit{RL policy} is based on \textit{PPO} algorithm \cite{b49} and trained by multiple phases. The experiment was run on a simple simulator with multiple scenarios, then real-world environments, and the results were positive. The policy can work up to 100 agents in a simple simulator, and 15 agents in high-collision-rate trajectories. In real-world scenarios, there are five real robots were tested in a lab, and one robot was run outdoors.

Different from the framework in \cite{b47}, \textit{PRIMAL} \cite{b46} built a decentralized multi-agent planner by combining reinforcement learning (RL) and imitation learning (IL). \textit{IL} gets expert's demonstrations from a traditional centralized motion planner, then use the demonstrations to train the \textit{RL} network which relies on the asynchronous advantage actor-critic (A3C) algorithm \cite{b48}. The \textit{RL} network's input is a limited field of views (FOV) of the environments around the robot's positions. After training, we have decentralized policies which can be transferred to any individual agents with various team sizes. The experiments were taken in grid-world environments and small workshop scenarios. On the grid-world environments, \textit{PRIMAL} could handle up to 1024 robots with the map size of $40 \times 40$. However, the paths' lengths from \textit{PRIMAL} are double the results in \cite{b47} and an underdog to the traditional motion planners. 
\subsection{\mrmp based on Robot Models}
Robot models have a significant affection on \mrmp problems by increasing exponentially workspace and complicating collision-checking algorithms. The robot models are grouped into three types: (1) points with no geometric features and no dynamics, (2) geometric shapes but no dynamics, and (3) robot models with geometric features and dynamics.

\subsubsection{Robots as Points with no Geometry and no Dynamics}
These problems are named \textit{MAPF} and require much less computational power comparing to other models. \textit{Search-based} algorithms presented in section \ref{sec:search_base} can solve these problems efficiently.

\subsubsection{Multi Robot Motion Planning with Geometric Constraints without Dynamics}

Considering robots with geometry, path plannings need to work on configuration spaces that are modified from workspaces by flattening techniques to handle the robot's geometry and orientation. 
The underlying idea is to build trees or graphs on the collision-free configuration spaces, then use them to find the robot paths. We can use both \textit{Search-base} algorithms or \textit{Sampling-based} algorithms to handle the problems.

In \textit{Search-based} algorithms, a $M^*$ variant \cite{b37}, \textit{CCBS} \cite{b39}, \textit{MC-CBS} are able to work on robot's models as shapes. A $M^*$ variant \cite{b37} use the principle of \textit{subdimensional expansion} method to apply to both \textit{RRT} and \textit{PRM} methods with considering the robot's shapes. Different to $M^*$, collision checking is extended to work with robot-robot shapes overlapping in the configuration space. 
In \cite{b38} and \cite{b39}, the authors leveraged \textit{CBS} framework to handle the geometric constraints in MAPF systems by adding more several constraints into constraint tree (\textit{CT}) nodes as expanding. \textit{Continuous Conflict Based Search - CCBS} \cite{b39} also uses a constraint tree \textit{CT} with two level searches, which is similar to \textit{CBS}. \textit{CCBS} works with robot geometries and continuous time by defining a new type of conflict: \textit{timed-actions} and a new constraint among time intervals to avoid the conflicts. The conflicts in \textit{CCBS} are defined as a tuple $<a_i,t_i,a_j, t_j>$ that means agent $i$ at time $t_i$ implementing action $a_i$ collide with agent $j$ at time $t_j$ doing action $a_j$. To avoid the collision, the first constraint can be imposed as $<i, a_i, [ t_i, t_i^u)>$ which stands for agent $i$ is not allowed to perform action $a_i$ from time $t_i$ to time $t_i^u$. Similarly, the second constraint is $<j,a_j, [ t_j, t_j^u)>$. Two new vertices are created, then each stores one of the constraints. To speed up the algorithm, \textit{CCBS} uses the concepts of \textit{cardinal}, \textit{semi-cardinal}, and \textit{non-cardinal} to decide whether making a new vertex or not and which vertex should be followed to expand the constraint tree. 

\textit{Multi-Constraint CBS} (\textit{MC-CBS}) \cite{b38} extended the constraint types in \cite{b39} to deal with rectangle, circle, or sphere agents, and named: mutually disjunctive constraint sets. Moreover, it uses heuristics to decide which node should be selected to expand. The \textit{CT} node's \textit{g-value} is a sum of all agents' costs from initial state to current state. The \textit{h-value} is the minimum vertex cost covering of a weighted conflict graph \cite{b38}. However, \textit{MC-CBS} handles well up to 6 agents.


Centralized approaches can provide a (probabilistic) completeness, however, do not scale well due to the curses of dimensionality. 
To enhance the scalability of PRM, Solovey \cite{b36} introduced an algorithm termed \textit{multi-robot discrete RRT} \textit{MRdRRT} which represents implicitly a composite roadmap as a tensor product of individual roadmaps in each configuration space. Instead of exploring all possible neighbor vertices, the algorithm discovers only a single vertex at each time step.


\subsubsection{MRMP with Geometry and Dynamics}

A challenge arising to motion planning with robot dynamics is planners need to work in state spaces and map the robots from the state spaces to configuration spaces to perform collision checking. There is a lot of effort working on this class, however, to the best of my knowledge, no algorithm can provide optimal solutions. We will discuss several non-complete algorithms and probabilistic complete algorithms.

Koenig \cite{b50} introduces a decoupled framework termed \textit{MAPF-POST} which uses a MAPF solver to determine a joint plan; then the joint plan is inputted to a temporal network to create a plan-execution schedule that can be executed on robots. The key components in \textit{MAPF-POST} are a data structure called Temporal Plan Graph (TPG) and a plan-execution schedule. The schedule assigns the times for robots to enter locations on the paths from the \textit{MAPF} solver. \textit{MAPF-POST} works well for a small number of robots. 
Yakovlev \cite{b41} proposed a fast and easy prioritized planner named \textit{Enhanced AA-SIPP(m)} which leverages \textit{AA-SIPP} framework \cite{b42}. This algorithm uses a re-assigning priority technique: assign the maximum priority to the robot which has failed to find a path, then replan for all robots. It can work with a large number of robots (up to 200 for some scenarios), however, the success rate reduces significantly in densely populated environments or complex environments. Running time is a huge negative point for this algorithm.
Both previous algorithms do not guarantee the completeness and cannot work in narrow corridor environments, because they don't take robot dynamics into account. It is difficult to apply those algorithms to real-world applications.

To handle huge state space, sampling-based motion planning could be the most appropriate algorithms for robot models with dynamics and geometry. The planners such as \textit{RRT-based} \cite{b6,b7,b11,b12}, \textit{KPIECE} \cite{b9}, \textit{CoSMMAS} \cite{b10} - discussed in the previous section \ref{subsec:samp-based} - approach \textit{MRMP} problem by centralized approach and building motion trees to find the paths. 

The most challenge in this class is how to expand robot's number in \mrmp systems. Due to exponential expansion of state space to the robot's number, all the algorithms struggle if there are more than 15 robots in the system. Reducing the state space dimensions is an active research direction in this class. Applying \textit{ML} as a guidance to expand the motion trees in \cite{b46,b47} shows potential results.

\subsection{Multi Robot Motion Planning based on Communication Type}
Communication is critical for multi-robot systems, especially as robots work in unknown or dynamic environments. As the robots exchange their data, the new data is crucial to motion planners to decide whether to replan or not. The robot's planners can be classified into two groups: fully-centralized and decentralized communication.

\subsubsection{Fully-centralized Approach}
In fully-centralized approaches, data from one robot can be transferred to all other robots synchronously. Each robot has all information that other robots possess. This setting is popular in various instances, especially when optimality or completeness is required in the performance. Popular applications are pickup and delivery, item retrieval in warehouse, etc. The algorithms discussed in section \ref{sec:search_base}, \ref{subsec:samp-based} are considered as fully-centralized communication, thus we  are not mentioned them further.

\subsubsection{Distributed Approach}

This approach assumes that each robot has no communication to other robots, or only exchanges data with neighbor ones, or communication can be interrupted sometimes. Under the assumption that robots are not able to provide optimal solutions, however, they can provide robustness and scalability, and are practical in real-world applications. A popular approach for \mrmp problems in this class is reactive methods, which replan the paths if there are updated data from the sensors or incoming messages.

Optimal reciprocal collision avoidance (ORCA) \cite{b59} is a fully decentralized algorithm in which robots are independent in making decision and does not communicate to each others. The collision avoidance relies on an idea that each robot has a fixed amount of time collision-free in the future. Therefore, each robot takes into account of observing other robot's velocities in order to avoid them. The robot has its velocity space, and not allow picking one on the forbidden area. The optimal velocity is calculated using linear programming. The algorithm can work well on thousand of robots in the simulation. Non-holonomic \textit{ORCA} \textit{(NH-ORCA)} \cite{b60} take dynamics and differential constraints into the robot models, and extend the algorithm into more realistic robot models. 

Asynchronous decentralized prioritized planning (ADPP) \cite{b58} works on a group of robots whose communication among the robots is asynchronous and is not guaranteed. By this setting, robot number can be changed during the mission. The algorithm uses a reactive approach in which the robots will check the \textit{consistency} of incoming messages via the data stored in data structure set \textit{Avoids}. If the \textit{consistency} fails, the robots will replan based on the new data. By doing so, the robot system can still operate even if a message is dropped.  

\mrmp with decentralized approach in communication has a huge potential because it avoids the exponential expansion of the state space. Improving the path quality is the open direction for this class of motion planning.
\subsection{\mrmp based on Environment Type}
\subsubsection{MRMP in Complete Known Environments}
Working with full information about environments is a well-research area for \textit{MRMP}. Most of the algorithms in the previous sections are used for this environment type, thus they are not discussed more here. We focus on algorithms that can guide robot systems to move efficiently in partially known environments.

\subsubsection{MRMP in Unknown Environments}

Unlike \mrmp problems in completely known environments, there is little interest so far in the setting where obstacles in robot workspaces are not initially known and are incrementally revealed online by robot's sensors. The existing approaches are constantly replanning as new objects are detected in the workspace. As updating the workspaces continuously, repeatedly representing the environments and robot replanning are two most expensive computations. 

Ayala \cite{b51} proposed a framework to work on GridWorld-based MAPF problems in Unknown environments. The input is expressed in terms of a co-safe linear temporal logic formula (scLTL), which must be satisfied by the derived paths. The major components are a weight finite deterministic transition system $\mathcal{TS}$, a deterministic finite automaton $\mathcal{A}$, and their weighted product automation $\mathcal{A^\mathcal{P}}$. $\mathcal{TS}$ is updated constantly its free-cell set, obstacle set, frontier set, and the connection weights between new cells and their neighbors as the agent's sensors detect new objects in environments; then $\mathcal{A^\mathcal{P}}$ and the trees are expanded. If the current paths are not accepted (not reaching goals yet), a frontier is selected by its weight $w$ to explore. The frontier's weights are the distance from the frontier to the robot's current position multiplying the promising-rate to find satisfied paths successfully from that frontier. 

Kantaros el at. \cite{b53} proposed a multi-robot reactive temporal logic Path Planning in unknown environments which are updated constantly by combining of sensor data. To reduce the runtime, which is proportional to the environment known portion, the temporal logic task is decomposed into a sequence of local point-to-point planning problems that are then solved online while adapting to the environment change. This planner is a complete algorithm. 

Guo at el. \cite{b61} introduces a knowledge transfer scheme for cooperative motion and task planning of multi-agent systems under local LTL specifications with partial known workspace. A proposed adjacency relation function incorporates the update data, and only being used if the corresponding states are marked “unvisited”. This technique can speed up the product automaton update.

Besides the regular \mrmp such as in \cite{b53}, in \cite{b52} the authors present a sampling-based approach that can synthesize perception-based control policies for multi-robot system in uncertain semantic environments. The algorithm incrementally builds trees based on exploring environments and reducing the uncertainty of the semantic label. This framework can handle with Integrated Task and Motion Planning mission. 

\mrmp in unknown environments are relative new research fields, whose scenarios are close to real-world applications. This class has a lot of potential to progress in the future.

\section{Open Problems} \label{sec:openpro}
Summarizing the ideas, algorithms, and frameworks from the previous sections, we can draw out several directions in \textit{MRMP} which still have room to develop.
\begin{itemize}
    \item Optimality: Multi-robot systems often require coordination to achieve goals efficiently. Although some algorithms \cite{b24, b25, b43, b39 } proved their optimality, however, they implemented under many assumptions or simplification on the robots or environments. Finding optimal plans for robots on real-world applications are challenging, especially when dealing with large numbers of robots with dynamics.
    \item Scalability: As robot's number in a multi-robot system increases, the problem becomes complex and exponentially expensive. \textit{PRIMAL} stated that in $40 \times 40$ grid-world environments, it can handle up to 1024 agents, however, the agents are just points with no dynamics or geometries. With geometries, \textit{CCBS} \cite{b39} can work up to 25 robots in some environments, and if considering dynamics, \cite{b10} can find plans efficiently up to 10 robots. Developing scalable algorithms for \textit{MRMP} problems remains a challenge.
    \item Real-time adaptability: To work in real-world applications, multi-robot systems must be able to adapt quickly to unexpected events or changes in the environment. Developing real-time adaptability algorithms that can handle unpredictable situations is an open problem.
    \item Heterogeneity: In most mentioned papers, the multi-robot systems are homogeneous, and the agent's capabilities are similar or identical. However, in real-world applications, multi-robot systems are heterogeneous with different sensing capabilities, speeds, and movement capabilities. Developing algorithms or frameworks that can handle such the heterogeneous systems is still an active area of research.
    \item Exploration and planning in unknown environments: multi-robot systems can be used in applications such as search and rescue, pollution, localizing wildfire, etc. In such the environments, motion planners need to continuously update their maps and find new efficient plans to guide robots exploring the dynamic environments while avoiding static and moving obstacles and preserving energy. 
\end{itemize}
\section{Conclusion} \label{sec:con}
In review, \mrmp is an active and rapidly developing field of research that has gained significant attention over the years. This survey presents a taxonomy overview, algorithms, and approaches that have been proposed and developed for multi-robot systems. 
We have discussed \mrmp classes such as robot model-based motion planning, planner-based motion planning, communication-based motion planning, and environment-based motion planning. 
Additionally, we have examined various categories of \mrmp methods, including centralized, decentralized, and prioritized approaches.

The study provides insights into the advantages and drawbacks of each algorithm and their applicability to specific scenarios and environments. In spite of significant progress in this field, there are still various open research problems that need to be addressed, such as optimality, scalability, real-time adaptability, improved performance, and robustness in dynamic and unknown environments.

In conclusion, \mrmp has a wide range of applications ranging from service robots to industrial assembly lines. With the ongoing advancement of technology, demand for efficient and reliable multi-robot systems will continue to grow. Thus, carrying-on research in this field to develop innovative algorithms and methods will play a crucial role to plan accurately motion of multiple robots to accomplish high-abstraction tasks and improve efficiency of various applications.



\vspace{12pt}
\color{red}

\end{document}